\documentclass{article}
\usepackage{INTERSPEECH2018,amsmath,graphicx,algpseudocode,algorithm,bm,cite,setspace,overpic}
\usepackage{subfig}
\usepackage{lipsum}
\title{
On Enhancing Speech Emotion Recognition using Generative Adversarial Networks}
\name{Saurabh Sahu$^1$, Rahul Gupta$^2$, Carol Espy-Wilson$^1$}
\address{$^1$Speech Communication Laboratory, University of Maryland, College Park, MD, USA \\
$^2$Amazon.com, USA
}  
\email{\{ssahu89,espy\}@umd.edu, gupra@amazon.com}

\begin{document}
\ninept
\maketitle
\begin{spacing}{0.92}
\begin{abstract}
Generative Adversarial Networks (GANs) have gained a lot of attention from machine learning community due to their ability to learn and mimic an input data distribution. GANs consist of a discriminator and a generator working in tandem playing a min-max game to learn a target underlying data distribution; when fed with data-points sampled from a simpler distribution (like uniform or Gaussian distribution). Once trained, they allow synthetic generation of examples sampled from the target distribution. We investigate the application of GANs to generate synthetic feature vectors used for speech emotion recognition. Specifically, we investigate two  set ups: (i) a vanilla GAN that learns the distribution of a lower dimensional representation of the actual higher dimensional feature vector and, (ii) a conditional GAN that learns the distribution of the higher dimensional feature vectors conditioned on the labels or the emotional class to which it belongs. As a potential practical application of these synthetically generated samples, we measure any improvement in a classifier's performance when the synthetic data is used along with real data for training. We perform cross validation analyses followed by a cross-corpus study.
\end{abstract}
%
%\begin{keywords}
%Generative adversarial nets, speech based emotion recognition
%\end{keywords}

\section{Introduction}
\label{sec:intro}

Emotion recognition has wide applications in psychology, medicine and designing human-computer interaction systems \cite{el2011survey}. 
In particular, using speech data for emotion recognition is popular because it's collection is easy, non-invasive and cheap. 
Given that datasets available for this task are typically limited in size, we explore synthetic feature generation and their utility for emotion recognition experiments. 
Generative adversarial networks (GANs) \cite{goodfellow2014generative} are popular tools that computer vision researchers have used to generate real looking synthetic images \cite{radford2015unsupervised} as well as for speech emotion recognition \cite{sahu2017adversarial, chang2017learning}.
We generate synthetic features to aid emotion classification using two schemes: (i) a vanilla GAN to generate a compressed version of the actual feature vectors and, (ii) a conditional GAN \cite{mirza2014conditional} to generate the actual higher dimensional feature vectors. 
The goal of our experiments is to assess the performance increase one can obtain with these synthetic features. 

GANs have enhanced state of the art in several tasks such as image generation \cite{wang2016generative}, image translation \cite{isola2017image} and dialog generation \cite{li2017adversarial}.
More recently, they have also been applied to the task of emotion recognition \cite{sahu2017adversarial,chang2017learning}.
However, these works have focused on learning feature representations for emotion recognition.
In this paper, we investigate the task of improving emotion classification accuracy using GANs. 
Initially, we train GAN models to imitate emotion utterance representations and generate synthetic samples. 
The synthetic datapoints are then used as features with/without real data and fed to a classifier. 
We observe increase in classification performances, indicating that even with only few hours of data, GANs can learn to generate synthetic samples learned on training data distribution. 
Finally, we do a cross validation study followed by cross-corpus experiments to obtain a more comprehensive assessment.

\section{Background: Generative Adversarial Networks}
%discrimimnator and generator error
%training procedure
%As said above different types. here we consider two types\\
% Simple 
% conditional\\
%Talk about simple architecture, training procedure, architecture picture, losses not converging for 1582. So, try a reduced dimension. Get reduced dimension using Adv auto. losses converge for 2D. \\
\begin{figure}[t]
\centering
\includegraphics[trim=0cm 0cm 0cm 0cm,clip=false,scale=.3]{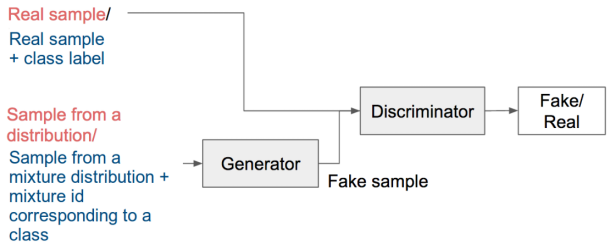}
%\captionsetup{justification=centering}
\caption{Block representation of a GAN architecture. A vanilla GAN requires access to real samples from a dataset and samples from a probability density (depicted in red font). A conditional GAN also requires the class samples corresponding to real datasamples and a mixture probability density (depicted in blue font). }
\label{fig:adv_auto}
\vspace{-6mm}
\end{figure}

%Talk about conditional, why conditional, training procedure and architecture. Encoder stronger than decoder why (references??). Ways to make the conditional GAN better -- 
A vanilla GAN consists of two components: a generator, $G$ and a discriminator, $D$.
Given a random sample $\bm z$ from a probability distribution $p_{\bm z}$, the generator is responsible for generating a fake datapoint $G(\bm z)$. 
The discriminator attempts to classify real samples $\bm x$ (drawn from a distribution $p_{\text{data}}$) against the one generated by the generator.
The objective of training a GAN is to obtain a generator that can mimic real data such that the discriminator is incapable of differentiating between real and fake samples.
GAN is trained using the following optimization on the GAN loss $V(D,G)$. 
\begin{equation}\label{eq:gan_loss}
\begin{aligned}
\min_{G} \max_{D} V(D,G) = \mathbb{E}_{\bm x \sim p_{\text{data}}} [\log D(\bm x)] + \\
\mathbb{E}_{\bm z \sim p_{\bm z}} [\log (1-D(G(\bm z)))]
\end{aligned}
\end{equation}
In the equation above, $D(\bm x)$ and $D(G(\bm z))$  are the probabilities that $\bm x$ and $G(\bm z)$ are inferred to be real sample by the discriminator.
Note that in the optimization in equation~\ref{eq:gan_loss}, the generator attempts to fool the discriminator as it tries to minimize $V(D,G)$. 
During GAN training, we minimize the discriminator and generator losses as defined below and track them separately. Note that for discriminator loss, $y$ is 1 if input is $\bm x$ and 0 if input is $G(\bm z)$.
\begin{equation}\label{eq:gan_loss2}
\begin{aligned}
&\text{\bf Disc. loss: } -y\log(D(\bm x))-(1-y)\log(1-D(G(\bm z))) \\
&\text{\bf Gen. loss: } -\log(D(G(\bm z))), \text{where } \bm x \sim p_{\text{data}} \text{, } \bm z \sim p_{\bm z}
\end{aligned}
\end{equation}
Although there are many variants of GAN architectures, we also experiment with a conditional GAN \cite{mirza2014conditional}.
Apart from the real data-points, conditional GAN also requires a class label for each data-point.
The distribution $p_{\bm z}$ is chosen to be a mixture distribution (e.g. Gaussian Mixture Models), where each mixture component corresponds to a sample class.
The objective of conditional GANs is to be able to generate fake samples for a class, when $\bm z$ is sampled from the corresponding component mixture in $p_{\bm z}$.
Figure~\ref{fig:adv_auto} provides a block diagram of vanilla/conditional GAN architectures.

\section{Synthetic Sample Generation for Emotion Recognition}
\label{sec:adv_emo}
%Emotion recognition from speech is a well-researched problem in which typically a classifier or regressor is trained on a set of extracted features. Usually these feature sets consist of hundreds or thousands of dimensions to capture the separability between various emotion classes. 
Training emotion recognition system often suffers from a lack of data availability.
As GANs have been successful in generating images, we explore their applicability in generating data samples for training emotion recognition systems.
Specifically, we focus on using vanilla and conditional GAN architectures to generate samples for each emotion class in our experiments and present our analysis.
As convergence of the loss $V(D,G)$ is often problematic, we also list the tricks we use to achieve the same.
We first describe the dataset we use for training the GAN models, followed by a description of sample generation strategy.

\subsection{Database for GAN training}
We use the Interactive Emotional Dyadic Motion Capture (IEMOCAP) dataset \cite{busso2008iemocap} for training GAN models. 
The dataset consists of five sessions of scripted and improvised interactions between two actors acting out real world situations. 
No two sessions have the same set of actors, enabling us to do a speaker independent leave-one-session-out five-fold cross validation. 
The database comes with the dyadic conversation segmented into utterances which are on an average about 5 seconds in duration. The utterances are then labeled by three annotators for emotion labels such as happy, sad, angry, excitement and, neutral (class labels are required for training conditional GANs).
We only use utterances for which we obtain a majority vote regarding the ground truth label.
Following \cite{kim2015emotion}, we combine the utterances in happy and excited class to get a ``combined happy'' class for our experiments. 
This was done to obtain a more balanced dataset, due to a small number of ``happy'' class instances.
For our classification experiments we focused on a set of 5531 utterances shared amongst four emotional labels: neutral (1708), angry (1103), sad (1084), and happy (1636). 
Overall, this amounts to approximately 7 hours of data.
%which was divided into train and validation splits for the cross-validation experiment.
 
We use the `emobase2010' feature set in openSMILE toolkit \cite{eyben2013recent} hat gives us a 1582-dimensional fixed length representation for each of the utterances.
It consists of several functionals computed from a set of acoustic low level descriptors \cite{eybenopen}. 
Next, we discuss the GAN training using these 1582-dimensional representations for each utterance.

\subsection{Sample generation using GAN}
Below, we describe the experiments performed using vanilla and conditional GANs separately.

\subsubsection{Sample generation using vanilla GAN}
\label{sec:van_gan}

%Not to mention the dataset available is generally limited in size. To that end we investigate the application of GANs to generate synthetic features to improve emotion recognition performance of classifiers. We start off with a cross-validation analysis followed by a cross-corpus experiment to study the generalizability of such approaches. Furthermore, the high dimensionality of features proves to be a problem for GAN convergence. We propose and discuss some methods to counter that. Specifically for each of the experiment settings we train two architectures of GANs: (i) A simple architecture with a generator and discriminator, where the generator's objective is to estimate a target probability distribution function (pdf) from a simpler pdf. (ii) A conditional GAN architecture where the generation of synthetic samples is conditioned upon the emotion labels which is also an input to the generator and discriminator.

In this experiment, we train a simple GAN architecture without the label information. 
Our initial aim was to generate synthetic 1582-dimensional feature vectors from a simple distribution $p_{\bm z}$ which was set as a 2 dimensional Gaussian distribution with zero mean and unit variance. 
Consequently, the generator is a feed-forward neural network with two neurons in the input layer and 1582 neurons in the output layer. 
Our discriminator is also a feed-forward neural network with 1582 neurons in input layer followed by two hidden layers and an output node with sigmoid activation. 
However, we could not get the generator and discriminator losses (equation \ref{eq:gan_loss2}) to converge. 
Any attempt towards changing the architecture, learning rates, number of epochs didn't lead to a convergence of losses. 
We hypothesized that this issue stems from a high dimensionality of feature space and the resulting data sparsity.
This prompted us to train a GAN to generate synthetic lower dimensional representations of the original higher dimensional representations. 

\begin{figure}[t]
\centering
%\captionsetup{justification=centering}
\begin{tabular}
{@{\hspace{-0.2cm}}c@{\hspace{0.0cm}}}
{\includegraphics[height=3.7cm, width=3.9cm]{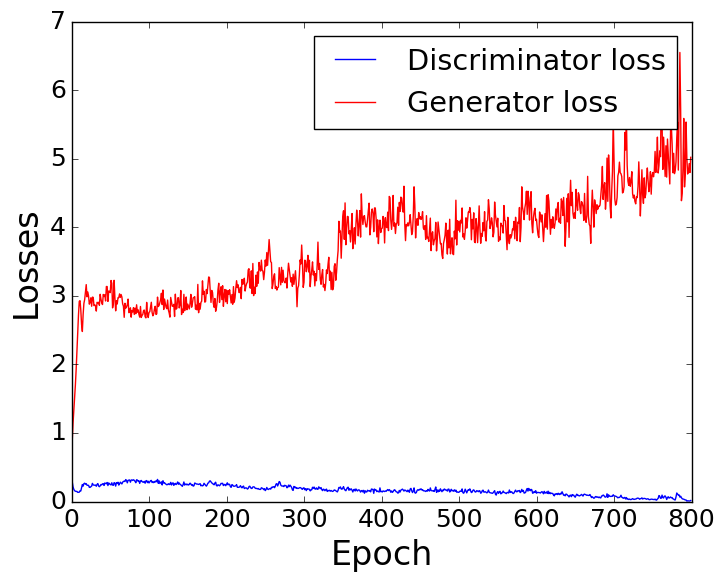}} 
{\includegraphics[height=3.7cm, width=3.9cm]{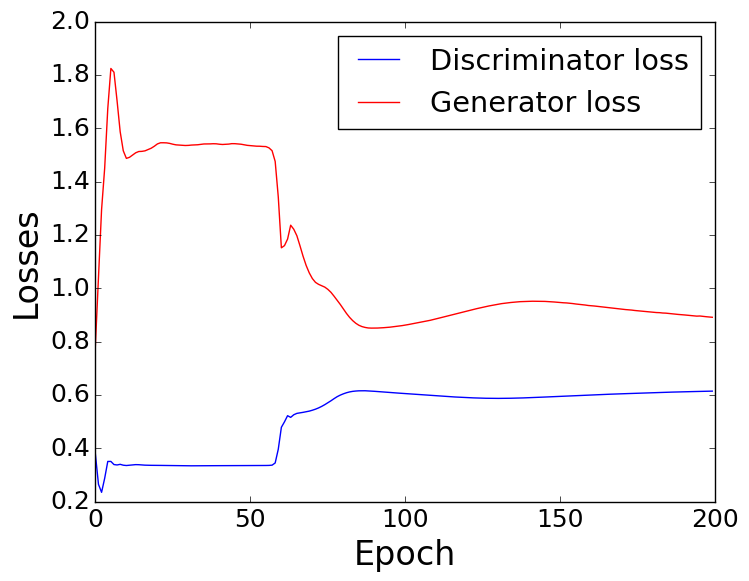}}
\end{tabular}
\vspace{-4mm}
\caption{Adversarial losses for GANs trying to estimate the actual high dimensional distribution (left) and their reduced 2-dimensional representation (right). Note how the errors can't converge while trying to estimate the higher dimensional distribution using a vanilla GAN.}
\label{fig:hyper}
\vspace{-2mm}
\end{figure}

\begin{figure}[t]
\centering
%\captionsetup{justification=centering}
%\begin{center}
\begin{tabular}
{@{\hspace{-0.3cm}}c@{\hspace{0.0cm}}c@{\hspace{0.0cm}}}
{\includegraphics[height=3.7cm, width=2.6cm]{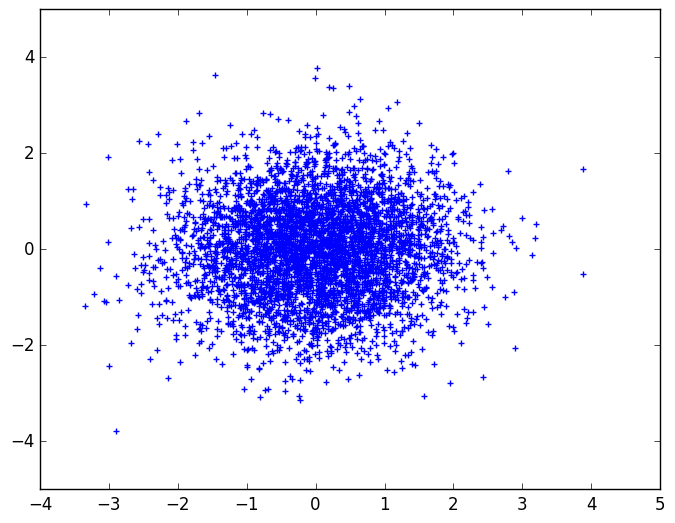}} 
{\includegraphics[height=3.7cm, width=2.6cm]{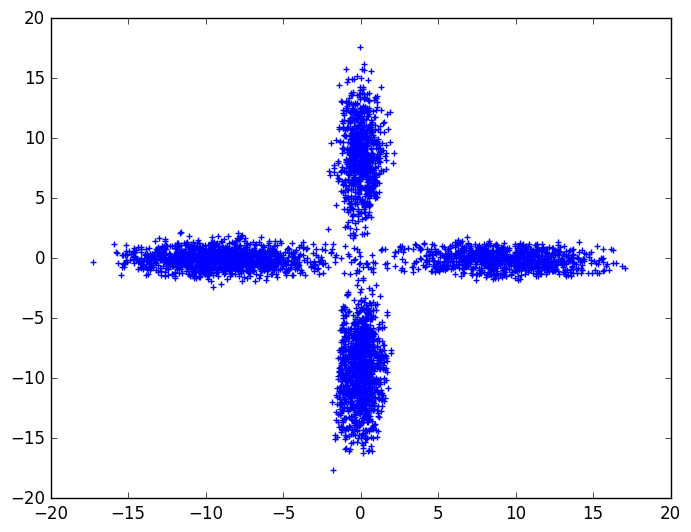}}
{\includegraphics[height=3.7cm, width=2.6cm]{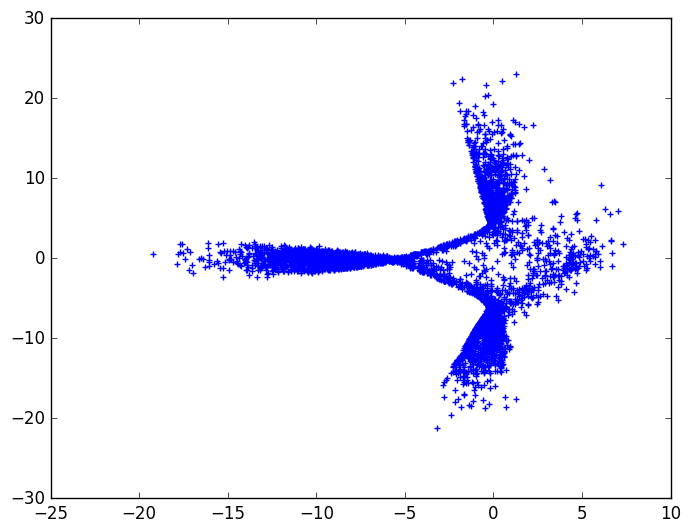}}
\end{tabular}
%\end{center}
\vspace{-3mm}
\caption{GAN was trained to transform a simple 2-D Gaussian distribution (left) to a lower-dimensional representation of the 1582-D feature vectors resembling mixture of four Gaussian components (center). After training, the generator's output distribution is shown on the right.}
\label{fig:simple_gan}
\vspace{-4mm}
\end{figure}

We use an adversarial auto-encoder framework \cite{sahu2017adversarial} to get the lower dimensional representations which has been shown to map the higher dimensional features onto a 2D space while preserving the cluster structure/relationship between feature vectors efficiently. 
The compressed feature representations resemble a GMM with four components, each GMM component corresponding to an emotion class \cite{sahu2017adversarial}. 
The output layer of generator now had two neurons and so does the input layer of discriminator. 
Input to generator are samples from a zero mean unit variance Gaussian distribution. 
Figure \ref{fig:hyper} shows the convergence of adversarial errors with the older and newer set up. 
While we had difficulty achieving convergence in synthetically generating 1582 dimensional feature vectors, vanilla GAN convergences when the target distribution was a 2 dimensional representation of original feature vectors.
%Once trained, the generator when fed with data points sampled from Gaussian distribution should be able to synthetically generate feature vectors that resemble the target distribution.
We generate synthetic data-points using the trained GAN and plot them in Figure~\ref{fig:simple_gan}.
We observe that we roughly obtain the four component GMM distribution. 

In our later experiments, we use the generated samples for classification.
Each synthetic data-point is assigned to an emotion class based on the GMM component yielding the highest membership for the generated data point.

\subsubsection{Sample generation using conditional GAN}
\label{sec:cond_gan}
We now focus on architectures that can generate synthetic higher dimensional feature vectors. 
We hypothesize that for a GAN to converge while trying to learn the distribution of higher dimensional representations, we need to provide it with more information. 
Conditional GAN is one such example where the synthetic data generation is conditioned on labels. 
Given a set of data-points $\bf{x} \sim p_{\text{data}}$ and their corresponding labels $\bf{y}$, a vanilla GAN models the distribution $\textbf{p(x)}$ while a conditional GAN learns the conditional distribution $\textbf{p(x{\textbar}y)}$. 
%Below we will demonstrate our methodology to get the adversarial errors of a conditional GAN to converge.
%For our baseline conditional GAN, we followed the architecture in Figure \ref{fig:adv_auto}. 
%It is quite similar to a simple GAN architecture in the sense that we are trying to learn a more complex target distribution from a simpler distribution except the fact that we are also providing it with label information. 
In our experiments, we chose $p_{\bm z}$ to be a mixture of four component GMM, with the target as modeling $p_{\text{data}}$ in the 1582 dimensional feature space. 
As is done typically for a conditional GAN, each mixture component in the GMM corresponds to a particular emotion. 
%We follow the same convention as \cite{sahu2017adversarial}. 
%For real Opensmile feature vectors, label information is provided to the discriminator by concatenating it with the corresponding one-hot label vector. 
%For data-points sampled from the mixture of Gaussians, we concatenate the generator's output with the one-hot vector corresponding to the mixture component before feeding it to discriminator. 
The class information for the real data-points as well as GMM components information during optimization are provided as one-hot encoded vectors.
We use several tricks to train a conditional GAN as described in detail below.

\begin{figure*}[t]
\centering
\includegraphics[scale=0.35]{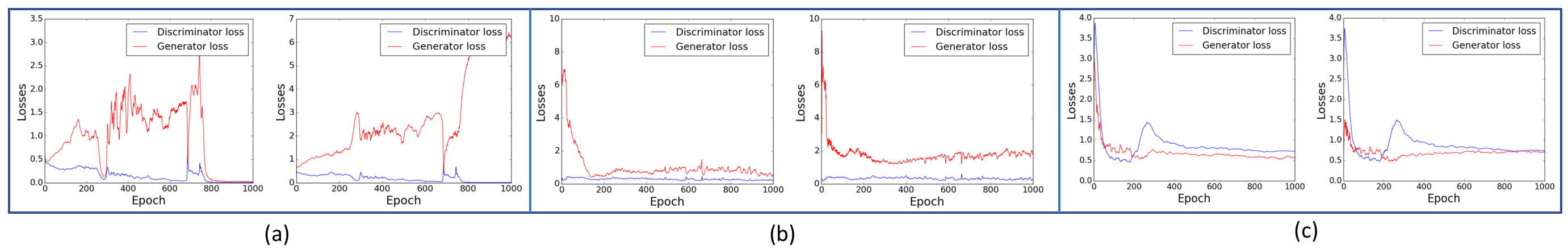}
\vspace{-8mm}
\caption{Convergence of generator and discriminator errors for training set (left) and validation sets (right) when (a) the generator of conditional GAN was randomly initialized (baseline-conditional), (b) the generator of conditional GAN was initialized using decoder's weights of a previously trained adversarial auto-encoder and, (c) along with weight initialization other schemes were used for better convergence (improved-conditional).}
\label{fig:cond_conv}
\vspace{-4mm}
\end{figure*}

First, we split the data-points in the IEMOCAP dataset into a training (4 sessions) and validation set (1 session).
For the baseline conditional GAN, we randomly initialize the generator and discriminator parameters. 
The learning rates of the generator and discriminator and the number of epochs for which they were trained are kept the same. 
Figure \ref{fig:cond_conv}(a) shows the plot of adversarial errors for the training and development splits, indicating a lack of convergence. 
Next we investigated the effect of initializing the network parameters based on a pre-trained network. 
We initialize the generator with decoder weights of a pre-trained adversarial auto-encoder (Figure 1 in \cite{sahu2017adversarial}). 
%Figure \ref{fig:decoder_cond} shows more stability the way the adversarial errors converge. 
%For both training and validation split, the adversarial errors seem to be converging.
Figure \ref{fig:cond_conv}(b) shows the plot of adversarial errors for the training and development splits.
We observe that while the losses converge both on the training and development sets, the discriminator error is very low.
This indicates that the discriminator is still able to distinguish between real data-points and the fake data produced by the generator.
%However, we notice that the discriminator's error is very low compared to generator indicating that it is having an easier time distinguishing between 'real' Opensmile feature vectors and the 'fake' data which is the generator's output. Furthermore, it doesn't seem to get any better with increasing the number of epochs. Hence, we need to come up with a way to efficiently teach the generator to mimic the distribution of real feature vectors. 
To improve the generator, we further incorporate two changes in our training scheme: (i) keeping the generator's learning rate higher than the discriminator (0.001 vs 0.0001 respectively) and, (ii) training the generator for five iterations for every iteration of discriminator training.
%We hope that with more weight updates, generator can learn better. 
Figure \ref{fig:cond_conv}(c) shows that this leads to a higher discriminator loss, indicating the generator is able to produce data-points that can fool the discriminator. 
%The low value of generator error suggests it is indeed able to fool the discriminator into thinking its output follows the pdf og real data. 
Training is not only stable but error convergence plots show that this training procedure also generalizes to the validation split. We refer to this model as improved conditional GAN.

We use the generated samples by the conditional GAN to improve emotion classification.
The synthetic samples generated are assigned a class based on the corresponding GMM component in $p_{\bm z}$.
In the next section, we describe our classification setup using the synthetically generated samples

\section{Classification using synthetically generated samples}
While the convergence of loss functions are a helpful tool to judge the capability of a trained GAN, we also investigate if the generated samples could aid emotion classification. To this end, we perform three sets of evaluations: (i) in-domain evaluation using synthetic samples as training set with and without real data (ii) in-domain evaluation on synthetic samples as test set and, (iii) a cross-corpus evaluation using a combination of real and synthetic data. For the simple GAN, we generate two dimensional representations of utterances to mimic the two dimensional representation learned by adversarial auto-encoders on real data. Additionally, we use the conditional GAN to generate 1582-dimension feature vectors to mimic the real data distribution.
The corresponding emotion classes for the data-points generated using vanilla and conditional are identified as specified in section~\ref{sec:van_gan} and \ref{sec:cond_gan}, respectively.
We train models to classify an utterance amongst the four emotion classes and use Unweighted Average Recall on test sets as our evaluation metric. 
We briefly describe each of these experiments below, followed by results using vanilla and conditional GAN.

\subsection{Synthetic samples in training set}
In this experiment, we perform a leave one session out cross-validation experiment on the IEMOCAP dataset.
Given that each session contains a unique pair of participants, this evaluation is also speaker independent.
We train the vanilla/conditional GAN on four IEMOCAP sessions and generate synthetic samples.
We train a Support Vector Machine (SVM) model with radial basis function as kernel, SVM$_{\text{van}}$, on the 2-dimensional projection space learned by the adversarial auto-encoder and train it under three conditions: (i) using only the synthetic samples generated by the vanilla GAN, (ii) using only the real samples in the four training sessions and, (iii) using a combination of both synthetic and real samples.
The trained models are evaluated on the 2-dimensions representations of the test set, as yielded by the adversarial auto-encoder.
Similarly, after obtaining samples from the conditional GAN, we train another SVM model, SVM$_{\text{con}}$, on the 1582-dimensional openSMILE feature space.
We again perform the three sets of experiments as mentioned above. SVM$_{\text{con}}$ is evaluated on the test partition in the 1582-dimensional openSMILE feature space.
The results for this experiment is listed in Table~\ref{tab:simple_gan_UAR}. It is clearly evident that by using only the synthetically generated samples for training the SVM we beat the chance accuracy by a big margin. It is worth noting that in case of simple GAN, the SVM’s performance trained with only synthetic data is comparable to that of a SVM trained with actual 2D code vectors. This could probably be because the 2D code vectors follow a specific distribution enforced by the adversarial auto-encoder framework and not just any random distribution. The specific distribution being the mixture of four Gaussian components is not as complex as real world distributions and hence the GAN model could easily learn that distribution. Furthermore, from Table~\ref{tab:simple_gan_UAR} it can be seen that while appending the real feature vectors with synthetic feature vectors from baseline conditional GAN can hurt the performance slightly that’s not the case when appending them with synthetic data points generated from improved conditional GAN. An improvement in UAR in both cases shows the potential of using synthetically generated data along with real data for training and classification purposes.

\subsection{Synthetic samples in test set}
In this experiment, in addition to using the real dataset to train the GANs, we also use them to train a SVM for emotion recognition. The synthetic samples were used in the test set. 
The objective of this experiment is to assess the similarity between real and synthetic data by using a model trained on real data to classify synthetic data. 
In case of vanilla GAN, the generated 2D representations were used as test set while the compressed 2D representations were used for training the SVM. 
For conditional GAN the higher dimensional feature vectors were used for training the SVM which was evaluated on the synthetically generated test set. 
Results are shown in Table~\ref{tab:synth_test}. As expected for the higher dimensional features the results shown are similar to what was obtained when the synthetic samples were used for training. 
For 2D samples the high accuracy suggests its much easier to estimate simpler dimensional distribution than a higher dimensional complex distribution.

\begin{table}[t]
\centering
%\captionsetup{justification=centering}
\caption {Classification results of a SVM trained using different set-ups involving the synthetically generated 2-D representation of the real 1582-D openSMILE and the synthetically generated higher dimensions samples with and without real data. Feature vectors of same dimensionality were used in the test set}
\vspace{-2mm}
\begin{tabular}{|l|c|} \hline
Dataset  & UAR (\%)  \\ \hline
Chance accuracy & 25.00\\
Only synthetic 2D code vectors & 55.38\\
Only 2D code vectors & 56.72\\
2D code vector + Synthetic  & {\bf 57.58}\\ 
Only improved-conditional & 34.09\\
Only real openSMILE & 59.42\\
Real openSMILE + baseline-conditional  & 59.20\\
Real openSMILE + improved-conditional  & {\bf 60.29}\\\hline
\end{tabular}
\vspace{-2mm}
\label{tab:simple_gan_UAR}
\end{table}

\begin{table}[t]
\centering
%\captionsetup{justification=centering}
\caption {Classification results of using synthetically generated vectors in the test set}
\vspace{-2mm}
\begin{tabular}{|l|c|} \hline
Dataset  & UAR (\%)  \\ \hline
2D code-vectors & 97.09\\
Improved-conditional & 35.23\\ \hline
\end{tabular}
\vspace{-6mm}
\label{tab:synth_test}
\end{table}

%\subsection{Synthetic samples in test set}
%In this experiment, in addition to using the real dataset to train the GANS, we also use them to train a Support Vector Machine (SVM) classifier for emotion recognition. The synthetic samples were used in the test set. In case of vanilla GAN, the generated 2D representations were used as test set while the compressed 2D representations were used for training the SVM. For conditional GAN the higher dimensional feature vectors were used for training the SVM which was evaluated on the synthetically generated test set. Results are shown in Table\ref{tab:synth_test}. As expected for the higher dimensional features the results shown are similar to what was obtained when only the synthetic samples were used for training. For 2D samples the high accuracy suggests its much easier to estimate simpler dimensional distribution than a higher dimensional complex distribution.
\begin{figure}[t]
\centering
%\captionsetup{justification=centering}
%\begin{center}
\begin{tabular}
{@{\hspace{-0.3cm}}c@{\hspace{0.0cm}}c@{\hspace{0.0cm}}}
{\includegraphics[height=3.5cm, width=4.2cm]{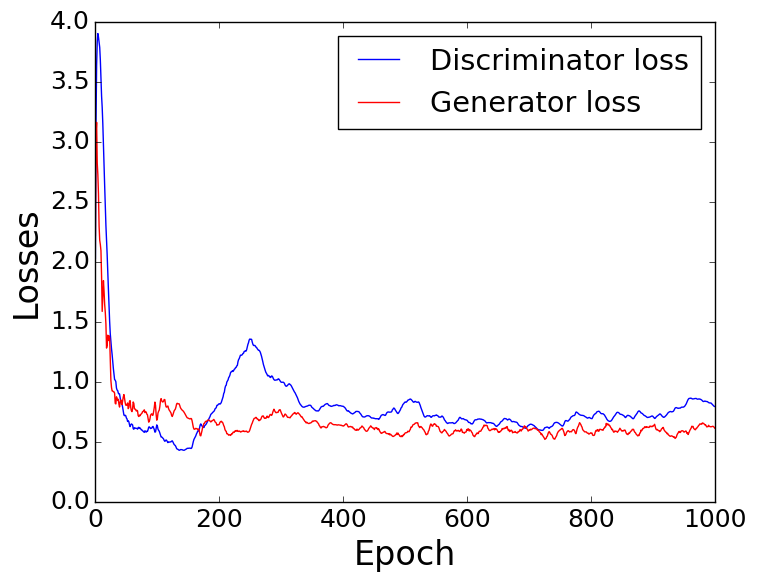}}
{\includegraphics[height=3.5cm, width=4.2cm]{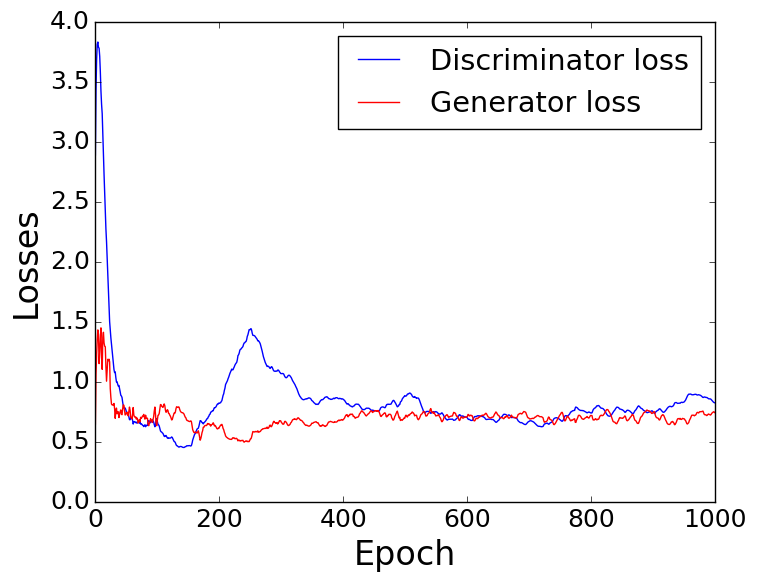}}
\end{tabular}
%\end{center}
\vspace{-3mm}
\caption{Loss curves showing the error convergence for a conditional GAN on training set (left) and test set (right) for cross-corpus experiments.}
\label{fig:cc_loss}
\vspace{-2mm}
\end{figure}
\subsection{Cross corpus experiments}
Having studied the convergence of GAN architectures and evaluating the quality of synthetically generated samples produced by them in a single corpora setting, we now move to performing cross-corpus evaluations.
The objective of this experiment is to investigate if synthetically generated samples can be used during classification on an external corpus (as opposed to being applicable for only in-domain tasks).  
We use IEMOCAP for training and MSP-IMPROV \cite{busso2017msp} as our  testing set. 
MSP-IMPROV, like IEMOCAP, also has actors participating in dyadic conversations which has then been segmented into utterances and annotated by evaluators. 
There are 7798 utterances in total spanned across the same four emotion classes.
However, the distribution across classes was highly unbalanced with the number of utterances belonging to happy/neutral class more than three times the number of angry/sad samples. 
This prompted us to use it as a test set rather than training set. The loss curves for a conditional GAN with the same set-up used in cross-validations experiments is shown in Figure~\ref{fig:cc_loss}.
We observe that the adversarial errors converge even if the test set is a different corpus than the training set. Results in Table~\ref{tab:CC_UAR} show a similar trend as cross-validation results.

%\begin{figure}[t]
%\centering
%\captionsetup{justification=centering}
%\begin{center}
%\begin{tabular}
%{@{\hspace{-0.3cm}}c@{\hspace{0.0cm}}c@{\hspace{0.0cm}}}
%{\includegraphics[height=3.5cm, width=4.2cm]{figures/train_mod__adv_loss_cc.png}}
%{\includegraphics[height=3.5cm, width=4.2cm]{figures/test_mod_adv_loss_cc.png}}
%\end{tabular}
%\end{center}
%\vspace{-3mm}
%\caption{Loss curves showing the error convergence for cross-corpus experiments. For a simple GAN learning the distribution of the lower dimensional representation (left) and for the conditional GAN trying to learn the distribution of training set (center) and test set (right) .}
%\label{fig:cc_loss}
%\vspace{-2mm}
%\end{figure}

\begin{table}[t]
\centering
%\captionsetup{justification=centering}
\caption {Classification results of a SVM trained on synthetic samples along with real training samples for different scenarios for cross-corpus experiments}
\vspace{-3mm}
\begin{tabular}{|l|c|} \hline
Dataset  & UAR (\%)  \\ \hline
Only synthetic 2D & 40.17 \\
Only 2D code vectors & 41.27 \\
2D code vector + Synthetic  & 41.54 \\ 
Only real openSMILE & 45.14\\
Only improved-conditional & 33.96\\
Real openSMILE + baseline-conditional  & 43.79\\
Real openSMILE + improved-conditional  & {\bf 45.40} \\\hline
\end{tabular}
\vspace{-5mm}
\label{tab:CC_UAR}
\end{table}

\section{Conclusions}
It is encouraging to observe that even with smaller datasets, the adversarial errors of a GAN can be made to converge. With more data it is expected that GANs will be able to learn a more generalized distribution/manifold where the openSMILE feature vectors lie. The experiments on conditional GAN show that a generator's job to estimate a more complex PDF from a simpler PDF is more complex than a discriminator's job which is to distinguish between fake and real samples. Hence, we had to incorporate tricks like updating the generator more times for a single update of discriminator or keeping the learning rate of generator more than that of a discriminator. We also experimented with reducing the number of trainable parameters in a discriminator but it didn't help in this case by a larger amount. While we see an improvement in performance of SVM when real data is appended with synthetic data, however the improvement isn't much. This is probably because the synthetic vectors after all are sampled from a distribution that mimics the real data distribution, something which the SVM classifier is already using for training. Also the smaller size of dataset might be hampering the capabilities of our GAN models. Cross corpus results showing similar trend as cross-validation indicate that the models are indeed generalizable across datasets with different priors.

In the future, we aim to further analyze other GAN architectures for the task of emotion classification \cite{goodfellow2016nips}.
A similar application of GANs could also be extended to other tasks within the study of emotion classification \cite{gupta2012classification}, as well as to tasks such as psychotherapy \cite{gupta2014predicting} and medicine \cite{gupta2016predicting}.

\end{spacing}
\newpage
\bibliographystyle{IEEEbib}
\bibliography{strings}

\begin{thebibliography}{10}

\bibitem{el2011survey}
M.~El~Ayadi, M.~S. Kamel, and F.~Karray,
\newblock ``Survey on speech emotion recognition: Features, classification
  schemes, and databases,''
\newblock {\em Pattern Recognition}, vol. 44, no. 3, pp. 572--587, 2011.

\bibitem{goodfellow2014generative}
I.~Goodfellow, J.~Pouget-Abadie, M.~Mirza, B.~Xu, D.~Warde-Farley, S.~Ozair,
  A.~Courville, and Y.~Bengio,
\newblock ``Generative adversarial nets,''
\newblock in {\em Advances in neural information processing systems}, 2014, pp.
  2672--2680.

\bibitem{radford2015unsupervised}
A.~Radford, L.~Metz, and S.~Chintala,
\newblock ``Unsupervised representation learning with deep convolutional
  generative adversarial networks,''
\newblock {\em arXiv preprint arXiv:1511.06434}, 2015.

\bibitem{sahu2017adversarial}
S.~Sahu, R.~Gupta, G.~Sivaraman, W.~AbdAlmageed, and C.~Espy-Wilson,
\newblock ``Adversarial auto-encoders for speech based emotion recognition,''
\newblock {\em Proc. Interspeech 2017}, pp. 1243--1247, 2017.

\bibitem{chang2017learning}
J.~Chang and S.~Scherer,
\newblock ``Learning representations of emotional speech with deep
  convolutional generative adversarial networks,''
\newblock in {\em Acoustics, Speech and Signal Processing (ICASSP), 2017 IEEE
  International Conference on}. IEEE, 2017, pp. 2746--2750.

\bibitem{mirza2014conditional}
M.~Mirza and S.~Osindero,
\newblock ``Conditional generative adversarial nets,''
\newblock {\em arXiv preprint arXiv:1411.1784}, 2014.

\bibitem{wang2016generative}
X.~Wang and A.~Gupta,
\newblock ``Generative image modeling using style and structure adversarial
  networks,''
\newblock in {\em European Conference on Computer Vision}. Springer, 2016, pp.
  318--335.

\bibitem{isola2017image}
P.~Isola, J.-Y. Zhu, T.~Zhou, and A.~A. Efros,
\newblock ``Image-to-image translation with conditional adversarial networks,''
\newblock {\em arXiv preprint}, 2017.

\bibitem{li2017adversarial}
J.~Li, W.~Monroe, T.~Shi, S.~Jean, A.~Ritter, and D.~Jurafsky,
\newblock ``Adversarial learning for neural dialogue generation,''
\newblock {\em arXiv preprint arXiv:1701.06547}, 2017.

\bibitem{busso2008iemocap}
C.~Busso, M.~Bulut, C.-C. Lee, A.~Kazemzadeh, E.~Mower, S.~Kim, J.~N. Chang,
  S.~Lee, and S.~S. Narayanan,
\newblock ``Iemocap: Interactive emotional dyadic motion capture database,''
\newblock {\em Language resources and evaluation}, vol. 42, no. 4, pp. 335,
  2008.

\bibitem{kim2015emotion}
Y.~Kim and E.~M. Provost,
\newblock ``Emotion recognition during speech using dynamics of multiple
  regions of the face,''
\newblock {\em ACM Transactions on Multimedia Computing, Communications, and
  Applications (TOMM)}, vol. 12, no. 1s, pp. 25, 2015.

\bibitem{eyben2013recent}
F.~Eyben, F.~Weninger, F.~Gross, and B.~Schuller,
\newblock ``Recent developments in opensmile, the munich open-source multimedia
  feature extractor,''
\newblock in {\em Proceedings of the 21st ACM international conference on
  Multimedia}. ACM, 2013, pp. 835--838.

\bibitem{eybenopen}
F.~Eyben, F.~Weninger, M.~W{\"o}llmer, and B.~Schuller,
\newblock ``open-source media interpretation by large feature-space
  extraction,''
\newblock .

\bibitem{busso2017msp}
C.~Busso, S.~Parthasarathy, A.~Burmania, M.~AbdelWahab, N.~Sadoughi, and E.~M.
  Provost,
\newblock ``Msp-improv: An acted corpus of dyadic interactions to study emotion
  perception,''
\newblock {\em IEEE Transactions on Affective Computing}, vol. 8, no. 1, pp.
  67--80, 2017.

\bibitem{goodfellow2016nips}
I.~Goodfellow,
\newblock ``Nips 2016 tutorial: Generative adversarial networks,''
\newblock {\em arXiv preprint arXiv:1701.00160}, 2016.

\bibitem{gupta2012classification}
R.~Gupta, C.-C. Lee, and S.~Narayanan,
\newblock ``Classification of emotional content of sighs in dyadic human
  interactions,''
\newblock in {\em Acoustics, Speech and Signal Processing (ICASSP), 2012 IEEE
  International Conference on}. IEEE, 2012, pp. 2265--2268.

\bibitem{gupta2014predicting}
R.~Gupta, P.~G. Georgiou, D.~C. Atkins, and S.~S. Narayanan,
\newblock ``Predicting client's inclination towards target behavior change in
  motivational interviewing and investigating the role of laughter,''
\newblock in {\em Fifteenth Annual Conference of the International Speech
  Communication Association}, 2014.

\bibitem{gupta2016predicting}
R.~Gupta and S.~S. Narayanan,
\newblock ``Predicting affective dimensions based on self assessed depression
  severity.,''
\newblock in {\em INTERSPEECH}, 2016, pp. 1427--1431.

\end{thebibliography}
\end{document}